\documentclass[runningheads]{llncs}
\usepackage[T1]{fontenc}
\usepackage{xcolor}
\usepackage{graphicx}
\usepackage[normalem]{ulem}

\usepackage[english]{babel}

\usepackage{amsmath}
\usepackage{amssymb}
\usepackage{graphicx}
\usepackage[colorlinks=true, allcolors=blue]{hyperref}
\usepackage{booktabs}
\usepackage{url}
\usepackage{subcaption}

\title{StyleX: A Trainable Metric for X-ray Style Distances}

\author{Dominik Eckert \inst{1}, Christopher Syben \inst{1}, Christian Hümmer 
\inst{1}, Ludwig Ritschl \inst{1}, Steffen Kappler \inst{1}, Sebastian Stober 
\inst{2}}

\institute{Siemens Healthineers, Germany\\
Otto-von-Guericke University Magdeburg, Germany\\}

\authorrunning{D. Eckert et al.}
\titlerunning{StyleX: A Trainable Metric for X-ray Style Distances}

\begin{document}
\maketitle

\begin{abstract}
The progression of X-ray technology introduces diverse image styles that need to be adapted to the preferences of radiologists. To support this task, we introduce a novel deep learning-based metric that quantifies style differences of non-matching image pairs. At the heart of our metric is an encoder capable of generating X-ray image style representations. This encoder is trained without any explicit knowledge of style distances by exploiting Simple Siamese learning. During inference, the style representations produced by the encoder are used to calculate a distance metric for non-matching image pairs. Our experiments investigate the proposed concept for a disclosed reproducible and a proprietary image processing pipeline along two dimensions: First, we use a t-distributed stochastic neighbor embedding (t-SNE) analysis to illustrate that the encoder outputs provide meaningful and discriminative style representations. Second, the proposed metric calculated from the encoder outputs is shown to quantify style distances for non-matching pairs in good alignment with the human perception. These results confirm that our proposed method is a promising technique to quantify style differences, which can be used for guided style selection as well as automatic optimization of image pipeline parameters.

\end{abstract}

\section{Introduction}


The process of acquiring X-ray images requires a detector to convert attenuated X-rays into a digital signal. 
However, the human eye cannot perceive the wide range of this signal at the same time and crucial diagnostic information is often contained in subtle changes~\cite{bruno2017256}.
Consequently, the signal must be compressed into a range of approximately 
500-1000 shades of gray \cite{barten1993spatiotemporal}. Due to the lossy nature of this 
compression, not all diagnostic information can be equally well preserved. This results in different 
styles of X-ray images, each with its own set of advantages.
Besides the actual presentation, the capability of the radiologists to extract the relevant 
diagnostic information depends on the training, personal preferences and neuro-physilogical processes. Therefore, radiologists have preferred X-ray image styles. 
Perceptual errors, which arise when radiologists fail to recognize the diagnostic information available in the image, account for 60-80\% of all diagnostic errors~\cite{waite2017interpretive}.
Nonetheless, radiologists encounter variations in X-ray machines, equipment modifications, and improvements in image processing pipelines, continuously.
These changes in style are 1) difficult to notice as most images show non-matching content 
and 2) may significantly influence the quality of their reading.
However, most of the medical images are processed by black-box vendor pipelines,
preventing the possibility to define distances between styles. In consequence an objective quantification is not possible.
Therefore, vendors subjectively identify the most suitable style in collaboration with the radiologist. 
This identification is based on example images from previous acquisitions. 
The aim is to align with the radiologist's preferred X-ray image style, thereby reducing the radiologist's effort to adapt to different image impressions. 
To overcome this subjective and manual process, we propose a deep learning-based style metric StyleX, that 
quantifies style differences. On the basis of style representations, the metric computes a distance between two stylized images,
reflecting the perceived style differences.
We propose a novel use of the Simple Siamese (SimSiam) approach~\cite{chen2021exploring} as a label-less unsupervised method
to learn style representations. With SimSiam, we bypass the need for explicit style distances labels in training and overcome
the need for content-matched pairs, a fundamental data limitation to construct a supervised deep-learning solution.

To our knowledge, no existing research in the field of medical imaging has specifically focused on 
developing a style metric for non-matching pairs. However, many studies have addressed the 
generalization of neural networks to inter-modality and intra-modality appearance differences, 
implicitly or explicitly addressing style differences \cite{li203domain,zhao2021mt,liu2021style}.
%
Most of these works rely on three different methodologies to address the style differences.
These methodologies might have either a style loss or style representations as a byproduct of their
training. 
First, GANs~\cite{paavilainen2021bridging,zhang2022c2,yang2019unsupervised,joshi2021nn}, which 
implicitly create a style loss by training the discriminator to distinguish between styles.
Armnaious et al.~\cite{armanious2020medgan} explicitly use an artistic style loss~\cite{gatys2016image} 
for PET-CT translation or \cite{hemon2023guiding} which utilizes an improved perceptual loss \cite{johnson2016perceptual}. Second, diffusion methods \cite{zhao2022egsde,kim2024adaptive}, which 
explicitly rely on a style loss to guide the inverse diffusion process, which often also happens to 
be a discriminator loss~\cite{ozbey2023unsupervised}. Third, methods using two encoders for disentanglement~\cite{gu2023cddsa,wagner2021structure} 
based on the work of~\cite{kotovenko2019content}, which incorporate an encoder in their training,
to generates style representations in the latent space.
However, in contrast our study uniquely develops a style metric, which is able to quantify style differences of 
non-matching pairs. 
In comparison, our proposed method does not depend on a decoder 
for embedding reconstruction, a pixel-wise loss, a discriminator or handcrafted style features. 
Moreover, our style loss is not merely a component to solve a larger task.
Rather, we specifically explore and refine the method for its capacity to 
accurately distinguish all styles.%

\section{Method}%
\subsection{StyleX - Metric}%
\begin{figure}[tb]
\centering
	\begin{subfigure}{0.49\textwidth}
		\centering
		\includegraphics[width=1.0\textwidth]{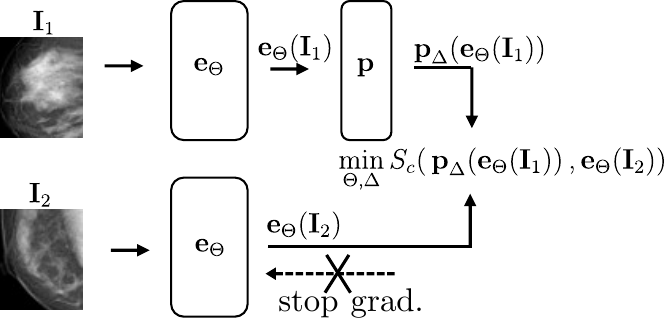}
	\end{subfigure}
	\begin{subfigure}{0.34\textwidth}
		\centering
		\includegraphics[width=1.0\textwidth]{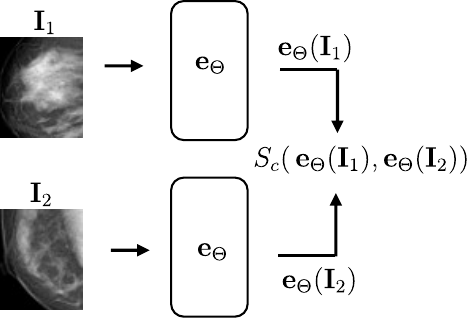}
	\end{subfigure}

	\caption{Training (left) of an encoder learning style representations and inference (right) to compute
    the StyleX distance between two images.}
	\label{fig:siamese_network}
\end{figure}

Our objective is to devise a style metric that enables quantifiable comparison between the 
styles of images of non-matching pairs.
For this, a multi-dimensional vector representing the style 
of an image $\mathbf{I}\in\mathbb{R}^{M\times N}$ is obtained using an encoder network 
$\mathbf{e_\Theta(\cdot)}\in\mathbb{R}^{M\times N}\mapsto\mathbb{R}^{D}$, with $M, N$ beeing the 
image dimensions, $D$ the embedding dimension and $ \mathbf{\Theta} $ the weights of the encoder. 
The distance between two vectors is then measured using cosine similarity $S_c$, which reflects the 
dot product between the vectors divided by each vector's Euclidean norm. We propose the metric StyleX, that 
can be formulated by:

\begin{equation}
	\mathrm{StyleX(\mathbf{I_1}, \mathbf{I_2})} = S_c(\mathbf{e_\Theta(\mathbf{I_1})} \cdot 
	\mathbf{e_\Theta(\mathbf{I_2})}) =
	\frac{\mathbf{e_\Theta(\mathbf{I_1})} \cdot 
	\mathbf{e_\Theta(\mathbf{I_2})}}{\|\mathbf{e_\Theta(\mathbf{I_1})}\| \cdot 
	\|\mathbf{e_\Theta}(\mathbf{I_2})\|},
\end{equation}
whereby $\mathbf{I_1}$,$\mathbf{I_2}$ are the images to compare. 
For a meaningful style metric, an encoder $\mathbf{e_\Theta}$ must be trained to generate style representations
that are disentangled from the content.
To be able to train the encoder on images from diverse sources 
we can not rely on parameters from imaging pipelines computing stylized
X-ray images, as they are either non-comparable or undisclosed vendor secrets, making them black boxes. 
Non-matching content also hinders pixel distance computation, like Euclidean distance.
Hence, without distance knowledge, no label exists, ruling out a supervised approach. 
However, to derive a meaningful distance metric, we need to discern not only which images share the same style and which do not, but also the degree of their difference. This contradiction rules out unsupervised methods, which 
depend on negative and positive pairs, like Siamese networks.
Consequently, to surmount these barriers, we propose employing the simple Siamese approach~\cite{chen2021exploring} (SimSiam) from Chen et al., a method originally developed 
to pre-train neural networks in an unsupervised manner to learn important image features for downstream tasks like segmentation or classification.
SimSiam relies solely on positive pairs, i.e., images sharing the same style. Chen et al. noted that it implicitly learns to embed distinct features, which allows to differentiate negative pairs, despite being trained only on positive pairs.
Hence, we propose to train the encoder in the SimSiam fashion, by presenting only matching images with the same style to the encoder.
Since no artificial distances must be enforced, this approach allows the style representations to be freely positioned in the embedding space.
We anticipate that the distances between these representations in the embedding space will reflect the stylistic differences between the images.
Following the idea of Chen et al., during training we utilize two asymmetries: 1) a neural network $\mathbf{p_\Delta(\cdot) \in\mathbb{R}^{D}\mapsto\mathbb{R}^{D}}$ with weights $\mathbf\Delta$, and 2)
a one-sided gradient flow, to prevent mode collapse. 
The training and inference setup is illustrated in Fig.~\ref{fig:siamese_network}.
For $\mathbf{e_\Theta}$, a ResNet-18 with an output embedding of $D=2048$ is used. $\mathbf{p_\Delta}$ and the hyper-parameters for the network and the training are as reported in~\cite{chen2021exploring}\footnote{\url{https://github.com/facebookresearch/simsiam}}.

\subsection{Raw Data}%
To train the encoder, we utilized publicly available digital mammograms (DM) from the Malmö Breast 
Tomosynthesis Screening Trial (MBTST)\footnote{\url{https://datahub.aida.scilifelab.se/10.23698/aida/mbtst-dm}}~\cite{zackrisson2018one,laang2016performance}.
The MBTST screened 14,851 women aged 40-74 using two-view digital mammography and one-view digital breast tomosynthesis at Skåne 
University Hospital, Malmö, Sweden. 
For our training and evaluation, we only used data where
breast density and thickness information was available, yielding a total of 7325 patients. Of these 7325 
DMs, we used 70\% (5064 images) for training and validation, and the remaining 30\% (2171 images) for testing.%
\subsection{Image Processing}%
We utilized two distinct image processing pipelines to create styles to validate our method: the
Linear Analysis Pipeline (LAP) for reproducible research and ease of analysis, and the
Proprietary Advanced Style System (PASS), an advanced closed-source prototype pipeline designed 
to process clinically relevant image impressions. LAP offers the advantage of simplicity due to its linear nature
and transparency as we fully disclose the pipeline in the supplementary material. It enables us 
to assess the network's capability to generate style representations with reasonable 
relationships with each other and allows us to associate these with the controlling parameters of LAP.
PASS, on the other hand, allows us to train and test our method on more complex and clinically relevant 
styles.

\textbf{LAP} outlines a straightforward Laplacian-pyramid-based imaging pipeline,
which weights the frequency bands and
maps a parameterized part of the full pixel range of the final image.
This pipeline has three adjustable parameters:
\begin{enumerate}
    \item \(w\) (window): defines the contrast of the image, by setting a pixel range, i.e., a window, which is mapped to the output image.
    \item \(l\) (low- to mid frequencies): determines how strongly medium-sized structures are highlighted or suppressed, by weighting low- to mid frequencies.
    \item \(h\) (high frequencies): emphasizes fine image details, by weighting high frequencies. 
\end{enumerate}
A detailed flowchart with the respective frequency bands and weightings is depicted in the supplementary material.



\textbf{PASS} is a black box with unknown underlying parameters. Therefore, the relationship between 
the produced styles is not known, but it is known whether images belong to the same style.
The pipeline is capable of generating 32 styles.
\textbf{Training:}
We process the MBTST training data using LAP and PASS to generate two distinct training sets. The encoder is then trained separately on each set.
We trained the encoder using the LAP pipeline, applying even values for parameters h, w, and l. For PASS pipeline training, we utilized 28 of the 32 available styles. For both training sessions, the images are initially cropped to dimensions of $800 \times 800$, and subsequently resized to $400 \times 400$. The training is conducted with hyperparameters reported in \cite{chen2021exploring}, with a batch size of 200.%

\newlength{\imwidth}
\setlength{\imwidth}{0.15\linewidth}

\setlength{\tabcolsep}{2pt} 

\begin{figure}[tb]
\centering
\begin{tabular}{ccccc}
\multicolumn{5}{l}{} \\

 $0.3 \cdot l_\text{max}  $ & $ 0.4 \cdot l_\text{max}$ & $ 0.5 \cdot l_\text{max}$ & $ 0.6 \cdot l_\text{max}$ & $ 0.7 \cdot l_\text{max}$ \\

	\includegraphics[width=\imwidth]{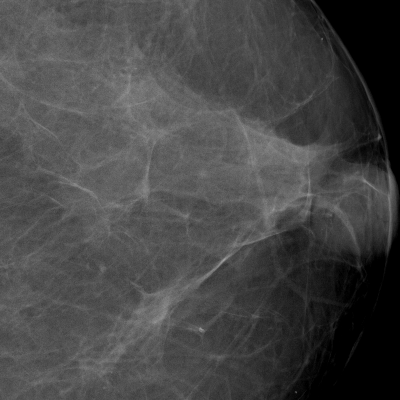} 
	&
	\includegraphics[width=\imwidth]{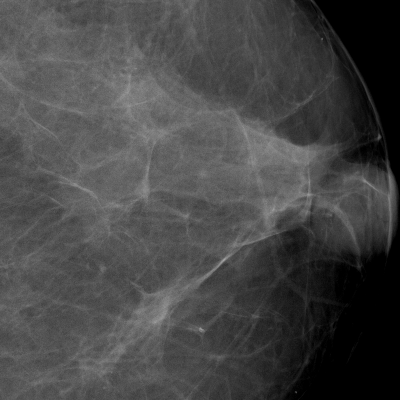} 
        &
 	\includegraphics[width=\imwidth]{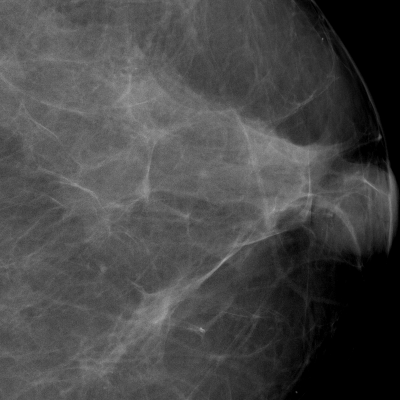} 
	&
	\includegraphics[width=\imwidth]{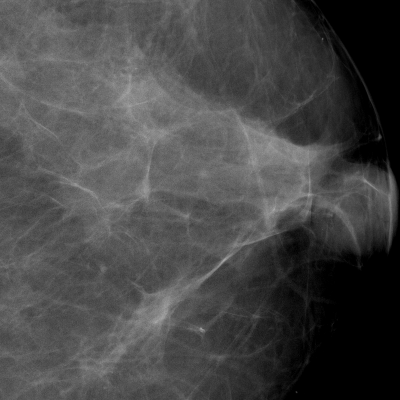} 
        &
 	\includegraphics[width=\imwidth]{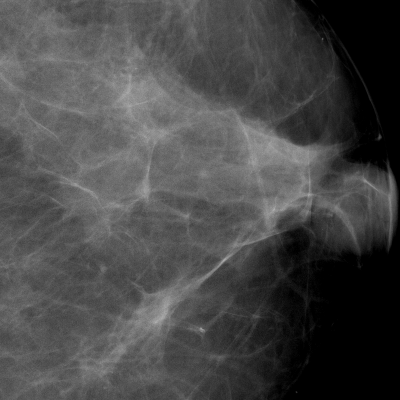} 
	\\	

 \hline
 \hline
 
\multicolumn{5}{l}{} \\
$w_\text{max} \, l_\text{min} \, h_\text{min} $  & $w_\text{min}\, l_\text{min} \, h_\text{max}$ & $w_\text{min}\, l_\text{max} \, h_\text{min}$ & $w_\text{min}\, l_\text{max} \, h_\text{max}$ & $w_\text{max}\, l_\text{max} \, h_\text{max}$ \\
	\includegraphics[width=\imwidth]{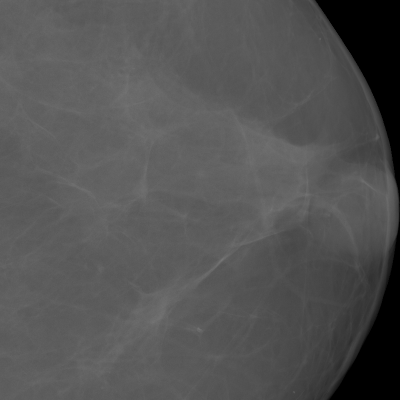} 
	&
	\includegraphics[width=\imwidth]{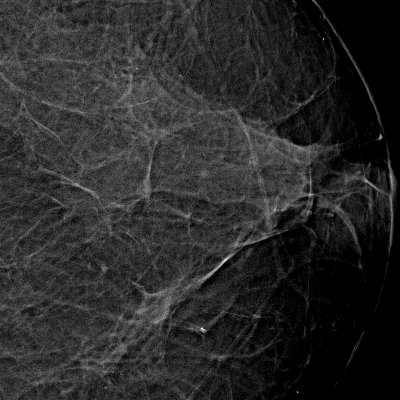} 
	&
	\includegraphics[width=\imwidth]{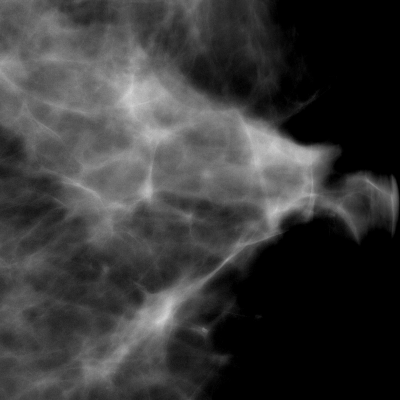} 
	&
	\includegraphics[width=\imwidth]{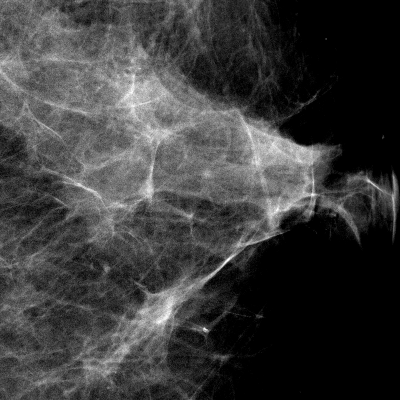} 
 	&
	\includegraphics[width=\imwidth]{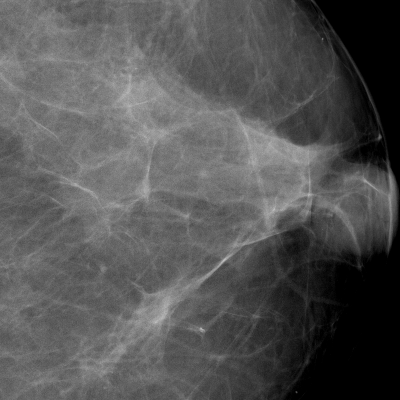} 
 \\
\end{tabular}
\caption{Top row: Styles generated with parameter sweep of middle frequencies.\\
Bottom row: Styles generated with max or min parameter settings.}
\label{fig_test_sets}
\end{figure}

\section{Experiments \& Results}

\begin{figure}[tb]
	\centering
	\begin{subfigure}{0.48\textwidth}
		\includegraphics[width=1.0\textwidth]{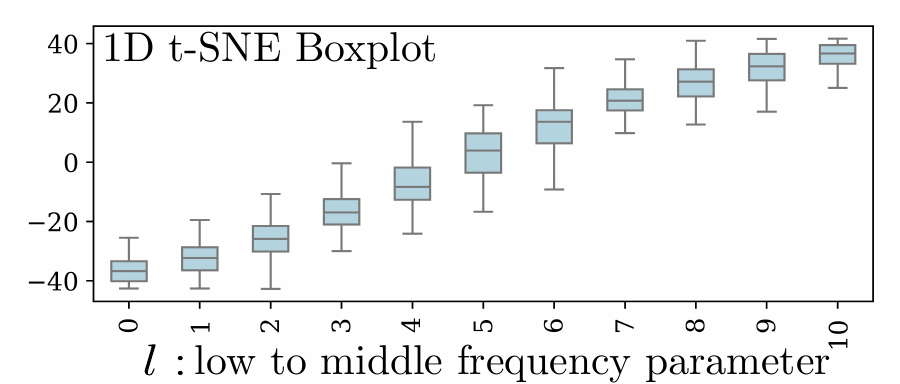}
		\caption{Parameter sweep for $l$ in LAP.}
        \label{fig:tsne1d1}
	\end{subfigure}
	\begin{subfigure}{0.48\textwidth}
		\includegraphics[width=1.0\textwidth]{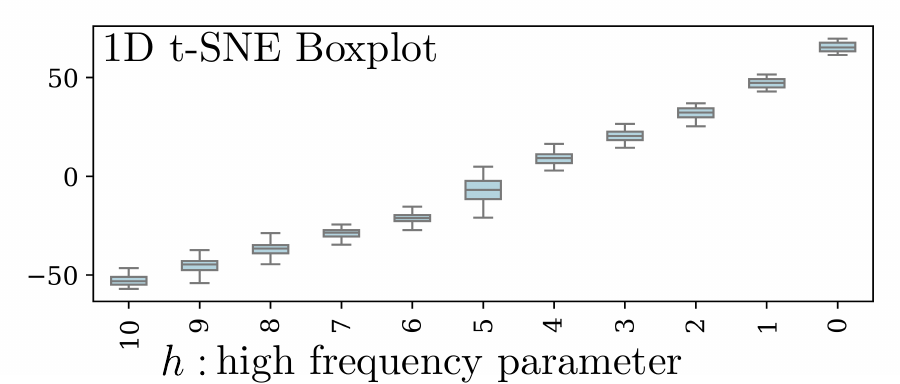}
		\caption{Parameter sweep for $h$ in LAP.}
        \label{fig:tsne1d2}
	\end{subfigure}

	\begin{subfigure}{0.48\textwidth}
		\includegraphics[width=1.0\textwidth]{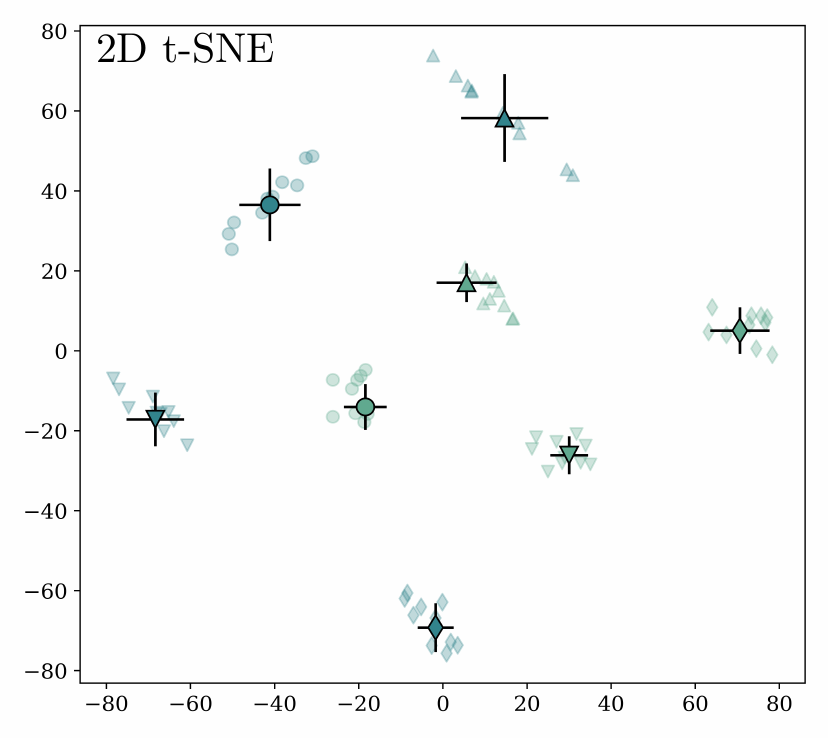}
		\caption{LAP Style representations.}
        \label{fig:tsne2d8}
	\end{subfigure}
	\begin{subfigure}{0.473\textwidth}
		\includegraphics[width=1.0\textwidth]{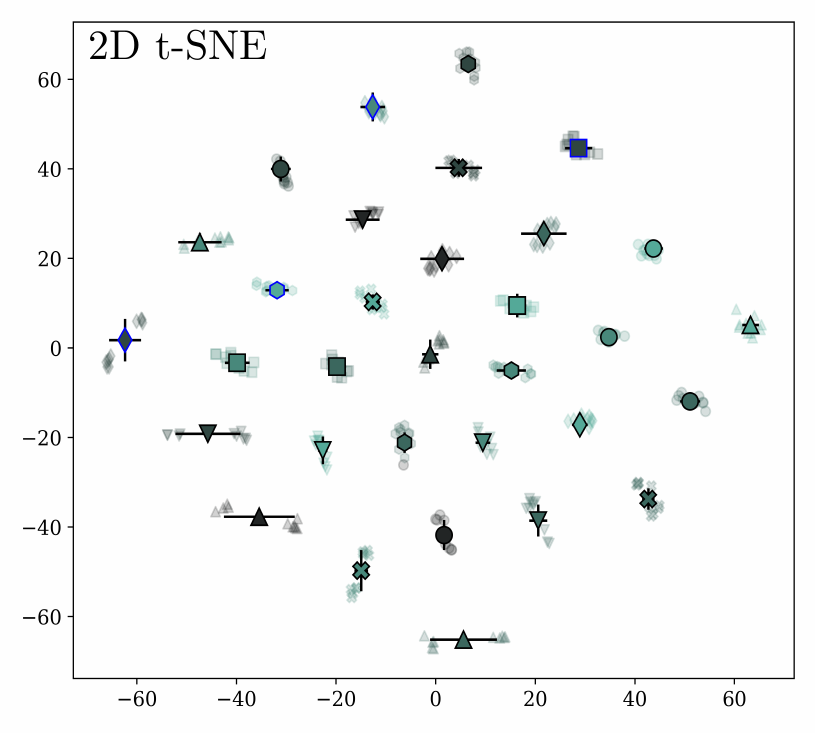}
		\caption{Pass Style representations.}
        \label{fig:tsne2d}
	\end{subfigure}

	\caption{Top: 1-D style representations of LAP-$l$ (\ref{fig:tsne1d1}) and LAP-$h$ (\ref{fig:tsne1d2}).
    Bottom: 2-D style representations of LAP-x (\ref{fig:tsne2d8}) and PASS-x (\ref{fig:tsne2d}).}

\end{figure}

To assess the performance of StyleX, we first investigate the ability of the encoder $\textbf{\textrm{e}}_\Theta$ to create meaningful and well-defined style representations. For a functional style metric, $\textbf{\textrm{e}}_\Theta$ must yield distinct style representations for different styles and similar ones for the same style. To be able to analyse the 2048 dimensional representations of the style representation, we can use t-SNE~\cite{maaten2009learning} to reduce the dimensionality to 1-D and 2-D, respectively. Note that, t-SNE attempts to preserve local neighborhood structure from the original high-dimensional space in the reduced low-dimensional space, thus allowing a visual investigation of the representations. 
A necessary property of the representations is that the magnitude of the distance between two representations must correlate with the degree of style difference. To investigate this behavior, we create three specialized test sets, using LAP, namely:
LAP-$w$, LAP-$l$ and LAP-$h$.
For each test set, we vary one of the three parameters of the LAP from its minimum to maximum value in 10 steps, while keeping the other two parameters fixed, resulting in images with subtle style changes. 
An example sweep of an image from the test set is depicted in the first row of Fig~\ref{fig_test_sets}. 
We then compute the style representations for each images of all three test sets with the encoder trained with the LAP training data and apply the 1-D t-SNE reduction to the representations. 
Given that only a single parameter among the three is altered at any given instance and considering the preservation of local neighborhoods by t-SNE, it becomes feasible to conduct an analysis of the correlation between the moving style parameter and the corresponding representations. 
The 1D representations are visualized in the boxplots of Fig.~\ref{fig:tsne1d1} and Fig.~\ref{fig:tsne1d2} for parameter sweep $l$ and $h$, respectively. 
Style representation created on images with identical parameter sets are depicted within the same box.
Furthermore, the style representations need to have high inter-cluster and small intra-cluster distances in the embedding space. To test this ability, we create a specialized test set using LAP, namely LAP-x. For this, the three parameters $w$, $l$ and $h$ are set to either their max or min values. This results in $3^2=8$ distinctive styles. Example images of the test set are shown in Fig.~\ref{fig_test_sets}.
The test set it processed by the encoder trained with the LAP training set and subsequently the 2-D t-SNE reduction is applied to the representations. 
The 2-D representations are visualized in Fig~\ref{fig:tsne2d8}. 
The variance of each cluster is marked with a line in each direction. 
Furthermore the two top-most outliers are visualized and 10 random points of each cluster.
To investigate the applicability of our proposed method to complex and clinical relevant styles, we generate a test set (PASS-x) with all 32 styles of PASS. 
As conducted on data processed with LAP, we reduce the computed representations using the 2-D t-SNE reduction. 
The 2-D representations for PASS-x are visualized in Fig.~\ref{fig:tsne2d}.

\begin{figure}[tb]
	\centering
	\includegraphics[width=0.8\textwidth]{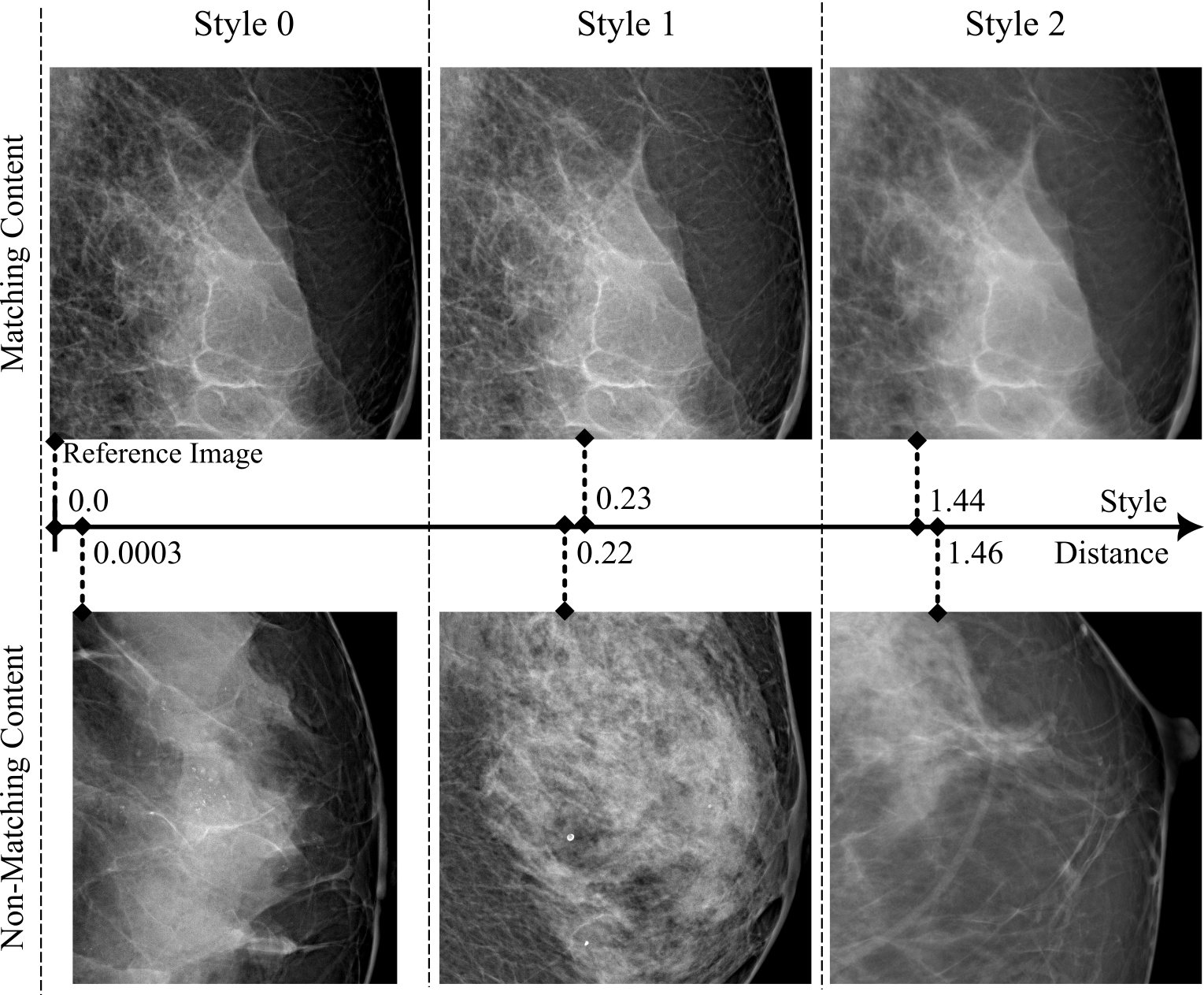}
	\caption{Example application of the StyleX. The style distance between all images and 
	the reference image at the top left of the figure is calculated. 
    The first row compares 
	images with different styles but same content as the reference image. 
    Images in the second 
	row have different content, and column-wise the same style.
	}

	\label{fig:example}
\end{figure}

Second, we investigate the the ability of StyleX to measure the distance between matching and non-matching pairs with complex and clinically relevant styles. 
For this, we apply StyleX to compute distances between a reference image and matching and non-matching pairs processed by PASS. 
The images with their computed distance are depicted in Fig.~\ref{fig:example}.

\section{Discussion}
The first two experiments are dedicated to analyze the capability of the encoder to learn and compute distinct style representations.
The knowledge of the white-box pipeline LAP enables us to analyze the style representation with respect to the parameters $l$, $h$ and $w$ of the imaging pipeline. 
The boxplots of the 1-D t-SNE reduction (cf. Fig~\ref{fig:tsne1d1},~\ref{fig:tsne1d2}) show that the style representations reflect the change of the moving parameter, i.e., increasing parameter $l$ or $h$ results in positional shift of the 1-D style localization. 
It is evident as the median values of the boxplots align with the ordering of the parameters. 
Note that styles generated with odd parameters were not seen during training, yet they are positioned correctly in the reduced 1-D space. 
This suggests that the encoder is
capable of interpolating to unseen styles. 
%
With the second experiment, the clustering behavior of the representations is analyzed. Both 2-D t-SNE plots exhibit distinctive clusters~(cf. Fig.~\ref{fig:tsne2d8}, \ref{fig:tsne2d}). This implies that the trained encoder is capable of creating style representations 
that are well-separated, provided that the styles being represented are distinctly unique. 
The representations of the clinically relevant images generated with PASS exhibit equally effective clustering as the style representations for images from LAP.
This indicates that our proposed encoder training strategy is capable of generating style representations of relevant clinical images without any further modifications.
Note that with PASS we have an increased number of 32 styles available, which does not present a challenge. 
Four of those 32 styles were not used during training, yet they are as well-clustered as the other styles. 
This suggests that the encoder has the ability to generalize to unseen styles. 
In conclusion, our study demonstrated the crucial property of StyleX’s encoder, which generates distinct representations positioned in space based on style differences.
Finally, we use StyleX to assess the correlation between the image impression and the computed distance, using both matching and non-matching content pairs (cf. Fig.~\ref{fig:example}). 
The visually perceived style differences between the images align well with the measured style distances. Moreover, the style 
quantification remains consistent whether comparing images with matching or non-matching content. 
This indicates that the measured style distance is not influenced by the image content.
In our research, we utilized StyleX on mammographic images, using the publicly accessible MBTST dataset of raw X-ray images.
However, this method is not exclusive to mammographic images. 
We are confident that StyleX’s application can be adapted to any X-ray image, regardless of the body region, and that the concept could even be extend to other modalities like MR. 
However, substantial style modifications necessitate encoder retraining.

\section{Conclusion}
This study introduces StyleX, a novel deep-learning metric for quantifying stylistic distances in medical X-ray images. 
We propose a novel use of the Simple Siamese (SimSiam) concept to train the encoder of StyleX, enabling computation of style representations. 
By exploiting SimSiam, we conduct training exclusively on identically styled pairs X-ray images, eliminating the need for predefined style distances and enabling compatibility with non-matching content. 
Our experiments confirmed our hypothesis: the encoder effectively produces distinct style representations with meaningful distances to each other, reflecting the perceived style differences. The applied metric successfully compared images with matching and non-matching content, matching the visual impression. Crucially, it quantifies style differences to the radiologists’ preferred style, even on a subtle level. This research establishes the foundation for a style loss in medical imaging, serving as an automatic style selector for radiologists or a loss function in imaging pipeline optimization.

\noindent\textbf{Disclaimer:} The concepts and information presented in this paper are based on research and are not commercially available.
\newpage

\bibliographystyle{plain}
\bibliography{sample}

\newpage
\appendix
\section{Appendix}
\begin{figure}
    \centering
    \includegraphics[width=1\linewidth]{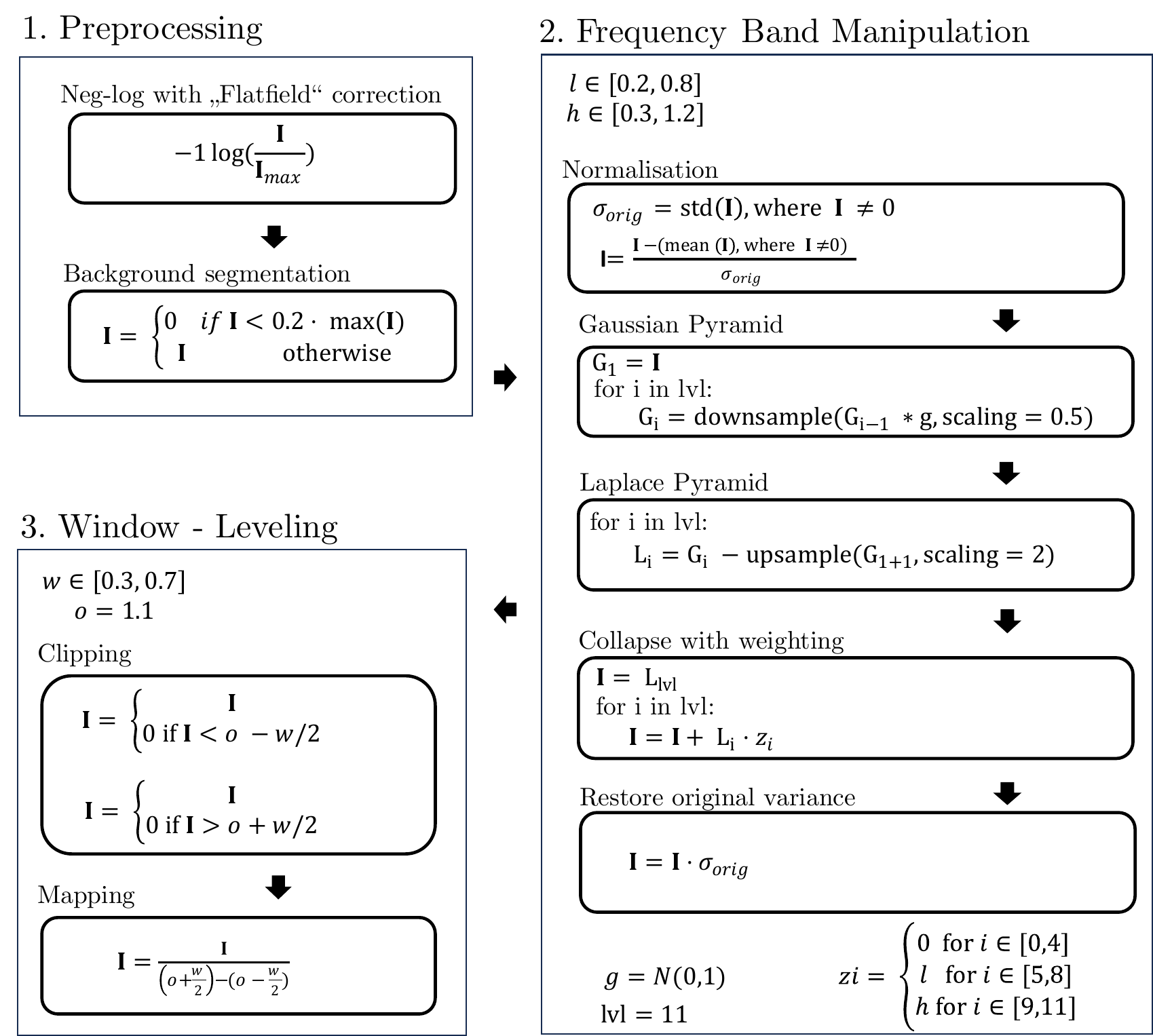}
    \caption{Description of the proposed LAP-Pipeline. Parameters $l$, $h$ and $w$ are applied within their defined ranges to produce X-ray image styles. In the paper, we refer to the actual parameter ranges from $h,w,l$ to range from 0 to 10.}
    \label{fig:enter-label}
\end{figure}

\end{document}